\documentclass[letterpaper, 10 pt, conference]{ieeeconf}  
\usepackage{xcolor}
\usepackage{booktabs}
\usepackage{multirow}
\usepackage[utf8]{inputenc}
\usepackage{siunitx}

\IEEEoverridecommandlockouts                              

\usepackage{amsmath}
\usepackage{amssymb}
\usepackage{siunitx}
\usepackage{graphicx}
\usepackage{gensymb}
\usepackage{cite}
\usepackage{graphicx}
\graphicspath{{imgs/}}
\usepackage{subcaption}
\usepackage{tikz}
\usetikzlibrary{shapes, arrows.meta, positioning, fit, backgrounds, calc}
\usepackage{xurl}
\usepackage[implicit=false]{hyperref}  
\hypersetup{
   breaklinks=true,   
   colorlinks=false,   
   pdfusetitle=false,  
}

\title{\LARGE \bf
Beyond Heuristics: A Standardized Real2Sim Pipeline for Physical Human Robot Interaction in Human-in-the-Loop Simulation}

\author{Chengyuan Yang$^{1}$, Yifan Wang$^{*1}$, Chun Kwang Tan$^{1}$, Sherwin Stephen Chan$^{1}$, Youlong Wang$^{1}$, \\ Xiaoyue Yan$^{1}$, Lei Li$^{2}$, Wei Tech Ang$^{1}$
\thanks{* Corresponding email: \tt\small ywang114@e.ntu.edu.sg}
\thanks{$^{1}$Chengyuan Yang, Yifan Wang, Chun Kwang Tan, Sherwin Stephen Chan, Youlong Wang, Xiaoyue Yan and Wei Tech Ang are with School of Mechanical and Aerospace Engineering, Nanyang Technological University, 639798 Singapore}%
\thanks{$^{2}$Lei Li is with Guangdong Zhongxin Intelligent Rehabilitation Research Institute, Foshan, China 528200}%
\thanks{Project page: \protect\href{https://jacekcoder.github.io/Bioweb2026/}{https://jacekcoder.github.io/Bioweb2026/}}%
}

\begin{document}

\maketitle
\thispagestyle{empty}
\pagestyle{empty}

\begin{abstract}
The aging global population drives demand for assistive robots, yet the safety risks and costs of physical testing make Human-in-the-Loop (HITL) simulation an attractive alternative. Its fidelity for coupled systems, however, is limited by interaction models whose impedance parameters are tuned heuristically rather than identified from data. We present a Real2Sim pipeline that identifies the coupled Physical Human-Robot Interaction (pHRI) dynamics of a pelvis--strap interface on an overground mobile balance assistant. The interface is modeled as a 6-DoF viscoelastic mechanism whose 12 directional stiffness and damping parameters are identified per subject via Covariance Matrix Adaptation Evolution Strategy (CMA-ES), using the user's ``Safe \& Comfortable'' feedback as a reproducible operating point that resolves harness-tightness ambiguity across anthropometrics. An intraclass-correlation analysis over a five-subject cohort separates shareable from subject-specific parameters, yielding a set of prior parameters derived from the existing data. Deploying this prior configures a previously unseen subject by refining only 5 of the 12 parameters. The calibrated model then reproduces the real interaction envelope and induces biomechanically accurate gait adaptations in the Human Digital Twin (HDT). Overly compliant and overly stiff settings, by contrast, fail as extreme settings, confirming a correct operating point that no heuristic tuning procedure can reliably select. The pipeline thus improves HITL simulation fidelity and supports the Human Digital Twin as a predictive tool for pre-clinical verification of personalized controllers.
\end{abstract}

\section{INTRODUCTION}
Global population aging is straining healthcare systems~\cite{abdi2019understanding}: the rising prevalence of neurological and musculoskeletal disorders degrades balance and mobility~\cite{rosso2013aging}, compromising independence~\cite{khan2022prediction}, while rehabilitation demand outstrips clinical capacity~\cite{jones2024healthcare}, motivating scalable technological interventions.

Assistive and rehabilitation robots \cite{peshkin2005kineassist,wang2023graceful,yan2025design,yang2025care,zhu2024you,gordon2024adaptable} offer substantial clinical benefit but are slow and costly to develop. Close physical contact imposes safety and ethical constraints: testing early-stage controllers on mobility-impaired users risks skin shear, joint hyper-extension, or loss of balance. At the same time, impairment heterogeneity \cite{moon2016gait} and users' adaptation to assistance \cite{beckerle2017human} demand extensive per-user personalization.

Human-in-the-loop (HITL) simulation \cite{ye2022rcare,erickson2020assistive,11127863,luo2024experiment,11247185} addresses this by enabling physical Human-Robot Interaction (pHRI) to be investigated safely in simulation, where fidelity hinges on how accurately the interaction is modeled. pHRI falls into two regimes: \emph{discrete} interactions with intermittent contact (e.g., feeding \cite{gordon2024adaptable}, dressing \cite{zhu2024you}), which introduce hybrid switching dynamics \cite{ye2022rcare,san2025simulating}; and \emph{coupled} interactions with continuous physical linkage (gait assistants \cite{peshkin2005kineassist,wang2023graceful}, exoskeletons \cite{yan2025design,yang2025care}), whose difficulty is dominated by the continuous, highly nonlinear constraints of compliant soft-to-soft interfaces (straps, cushions). This coupled regime is the one addressed here.

Current HITL simulations of coupled interaction rely on spring-damper or impedance proxies. This 6-DoF representation, with direction-specific linear and rotational stiffness and damping, is itself well established. Luo et al.~\cite{luo2024experiment}, for example, couple an exoskeleton to its wearer through per-axis bushing elements. What remains heuristic is how the constants are \emph{obtained}: even state-of-the-art works~\cite{11127863,luo2024experiment} assign them by empirical testing or visual inspection, without grounding them in real coupled motion data or assessing inter-subject variability. This is compounded by harness tightness, also set empirically to trade off ``safety'' and ``comfort,'' which obscures any single ``correct'' nominal model. Moreover, no prior work tests whether such models induce \emph{biomechanically correct} responses in the digital twin. We therefore propose a standardized Real2Sim pipeline answering three questions: \textbf{(RQ1)} how can qualitative tightness (``Safe'' vs.\ ``Comfortable'') be quantitatively defined, and how does it alter coupling dynamics; \textbf{(RQ2)} do identified pHRI parameters share common cross-subject distributions usable as generalized priors; and \textbf{(RQ3)} how can we verify that the model induces biomechanically accurate responses in the digital twin?

To answer these, we present a Real2Sim pipeline for gait-assistive HITL simulation. The pelvis--strap interface is formulated as a 6-DoF viscoelastic mechanism embedded in a full multibody simulator, whose 12 directional stiffness and damping parameters are identified per subject from real coupled motion via CMA-ES, so directional anisotropy emerges from data rather than being assigned~\cite{11127863,luo2024experiment}. Data collection is anchored to the ``Safe \& Comfortable'' operating point, and an intraclass-correlation analysis across five subjects separates shareable from subject-specific parameters into deployable priors. Validated on an unseen subject, the calibrated model reduces Real2Sim discrepancy relative to extreme isotropic baselines and induces biomechanically accurate adaptation in the digital twin.

\section{OVERVIEW OF THE HUMAN-IN-THE-LOOP SIMULATION FRAMEWORK}
\begin{figure}[htbp]
    \centering
    \makebox[\linewidth][c]{\includegraphics[width=\linewidth]{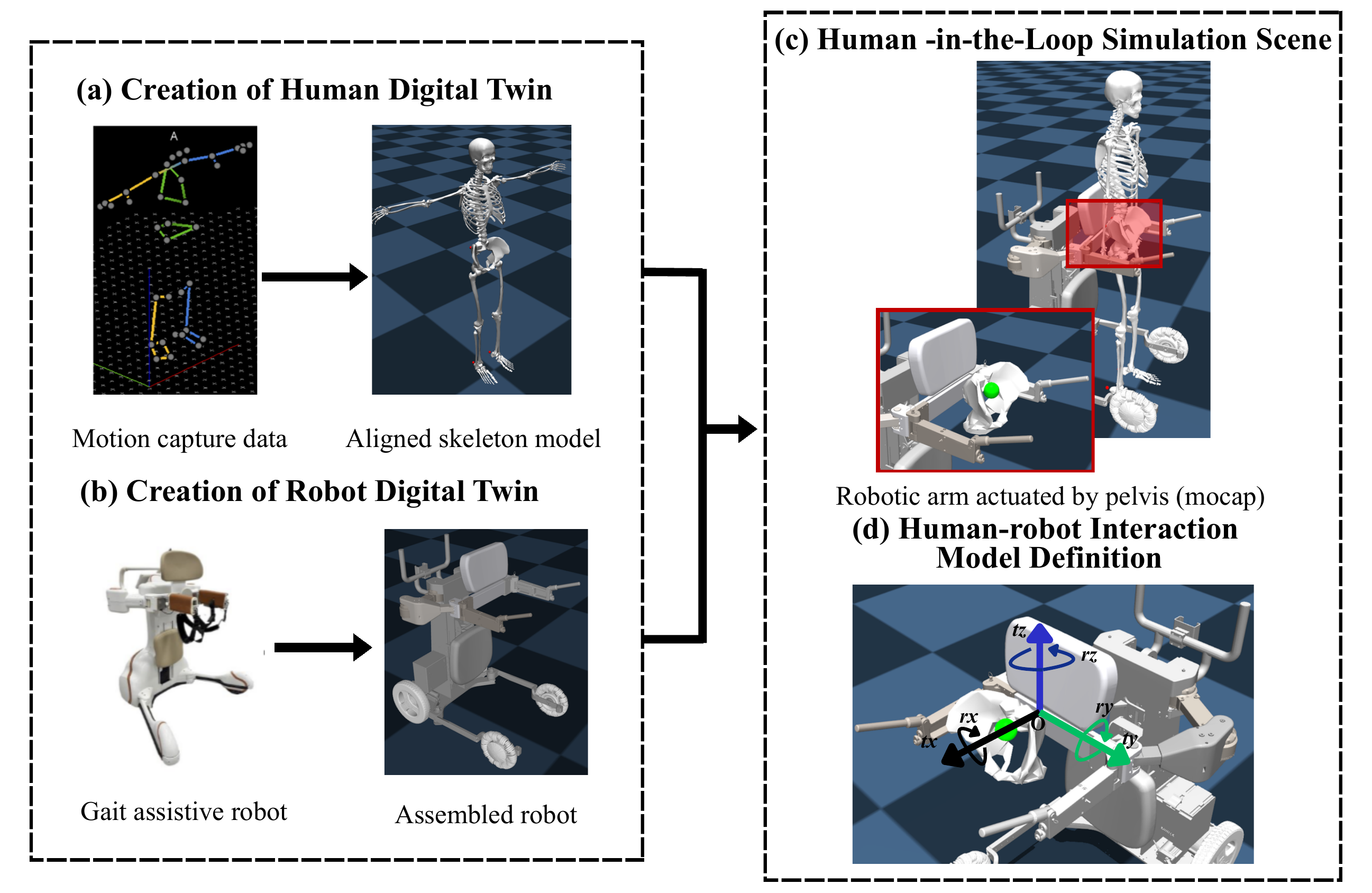}}
    \caption{Overview of the HITL simulation environment. The Human Digital Twin, robot digital twin, and pelvis--interface coupling are modeled in MuJoCo.}
    \label{fig:hitl_overview}
\end{figure}

This work builds on our established HITL framework~\cite{11127863} (Fig.~\ref{fig:hitl_overview}), which co-simulates human and robot dynamics. The Human Digital Twin (HDT) is a 27-DoF full-body skeletal model driven by a deep reinforcement-learning (RL) policy to emulate naturalistic gait; the robot digital twin replicates the Dynamic Robotic Balance Assistant (DRBA) \cite{wang2026drba}, following the user from behind and intervening on balance loss.

The pelvis--interface coupling is modeled as a 6-DoF viscoelastic mechanism: a virtual 6-DoF joint (three prismatic, three revolute) connects the pelvis frame to the interface frame, with generalized coordinates $\mathbf{q}(t)=
\big[t_x,\ t_y,\ t_z,\ r_x,\ r_y,\ r_z\big]^{\mathsf T}\in\mathbb{R}^6$ encoding their relative translation and rotation and conjugate interaction wrench $\mathbf{F}_{\mathrm{int}}(t)= \big[F_x,\ F_y,\ F_z,\ \tau_x,\ \tau_y,\ \tau_z\big]^{\mathsf T}\in\mathbb{R}^6$. Within the joint limits, the wrench follows a spatial spring--damper law with both linear and rotational/torsional properties:
\begin{equation}
\mathbf{F}_{\mathrm{int}}(\mathbf{q},\dot{\mathbf{q}})
=
\mathbf{K}\,\mathbf{q}
+
\mathbf{D}\,\dot{\mathbf{q}},
\qquad
\mathbf{q}_{\min}<\mathbf{q}<\mathbf{q}_{\max},
\label{eq:restricted_msd}
\end{equation}
with stiffness $\mathbf{K}=\mathrm{diag}(\mathbf{k})$ and damping $\mathbf{D}=\mathrm{diag}(\mathbf{d})$; when $\mathbf{q}$ reaches the limits, rigid
constraint forces/torques are generated to prevent exceeding the predefined bounds. Consequently, the interaction characteristics are governed by the tuning of $\mathbf{K}$, $\mathbf{D}$, and the joint limits $\mathbf{q}_{\min}$, $\mathbf{q}_{\max}$.

\section{APPROACH}
The proposed Real2Sim pipeline comprises four stages: (1) \textit{Tightness Quantification}, normalizing subjective harness feedback into a reproducible operating point; (2) \textit{Two-Stage Calibration}, first fixing the joint limits and then identifying the viscoelastic parameters; (3) \textit{Shareability Characterization}, separating shareable from subject-specific parameters into deployable priors; and (4) \textit{Model Validation} of the resulting model.

\subsection{Tightness Quantification and Normalization}\label{sec:tightness}
Harness tightness is ambiguous: an identical drawn-out strap length $T$ yields very different pressures across anthropometrics, precluding direct comparison. We therefore normalize $T$ about a subject-specific psychophysical operating point:
\begin{equation}
    \tilde{T} = \frac{T - T^*}{(T_{\max}-T_{\min})/2},
    \label{eq:method4}
\end{equation}
where $T$ is a discrete ordinal level (1-cm increments of remaining strap length at the buckle), $T_{\max}$ the tightest tolerable and $T_{\min}$ the loosest constraining level, and $T^*$ the subject's self-selected ``Safe \& Comfortable'' level. To standardize this selection, participants performed a ``crouch'' test verifying the setting constrained vertical descent without slippage; $T^*$ is thus the most comfortable setting that ensures safety without excessive abdominal restriction.

\subsection{Two-Stage Calibration}\label{sec:two-stage calibration}
The identification methodology proceeds in two sequential stages: (1) calibration of the joint limits $\mathbf{q}_{\min},\mathbf{q}_{\max}$, and (2) dynamic optimization of the viscoelastic parameters.

First, the joint limit parameters $\mathbf{q}_{\min}$ and $\mathbf{q}_{\max}$ are calibrated. Participants perform maximum-range pelvic motions along each of the six coordinates of $\mathbf{q}$ (the experimental procedure is detailed in Sec.~\ref{sec4:experiment}). For every coordinate $i \in \{t_x, t_y, t_z, r_x, r_y, r_z\}$, the minimum and maximum values recorded during these motions are taken as the corresponding entries of $\mathbf{q}_{\min}$ and $\mathbf{q}_{\max}$, thereby defining the natural ``slack'' range allowed by the strap before physical engagement occurs. These calibrated bounds are then used directly as the joint limits $\mathbf{q}_{\min}$, $\mathbf{q}_{\max}$ of the spring--damper model in Eq.~\ref{eq:restricted_msd}, delimiting the viscoelastic regime within which the interaction wrench acts before the rigid constraint forces engage in simulation.

Subsequently, with the joint limits fixed, $\mathbf{K}$ and $\mathbf{D}$ are identified by minimizing the discrepancy between measured and simulated interface motion. We track the two strap--pelvis attachment points $c\in\{L,R\}$ individually rather than collapsing them to a midpoint, because averaging cancels the antiphase component that encodes pelvic rotation; we also penalize velocity residuals so the damping terms are excited by the dynamics. With optimization vector $\theta=[\mathrm{diag}(\mathbf{K})^\top,\mathrm{diag}(\mathbf{D})^\top]^\top$, position residual $\mathbf{e}_p^{c}=\mathbf{p}_{\mathrm{sim}}^{c}(\theta)-\mathbf{p}_{\mathrm{meas}}^{c}$ and its time derivative $\mathbf{e}_v^{c}$, and pooling $\langle\cdot\rangle$ over all samples and both sides, the per-trial cost non-dimensionalizes the squared residuals by characteristic scales $\sigma_p=41$\,mm, $\sigma_v=80$\,mm/s:
\begin{equation}
J_d(\theta)=\frac{\big\langle \lVert\mathbf{e}_p^{c}\rVert^2\big\rangle}{\sigma_p^2}+\frac{\big\langle \lVert\mathbf{e}_v^{c}\rVert^2\big\rangle}{\sigma_v^2},
\end{equation}
and the full objective minimizes the mean of $J_d$ over the trials, yielding the calibrated parameters $\theta_{\mathrm{Opt}}=(\mathbf{K}^\ast,\mathbf{D}^\ast)$.


To solve this non-linear, non-convex problem, we employ CMA-ES~\cite{hansen2016cma} for its robustness on discontinuous, constraint-induced search spaces. Because the parameters span several orders of magnitude, the search runs in log-space $z=\ln\theta$, rendering the step size scale-invariant, within physically admissible bounds derived from strap mechanics.\footnotemark
\footnotetext{\emph{Search bounds} $k_t\in[150,8000]$\,N/m (up to $10^4$ on $t_z$); $k_r\in[15,500]$\,Nm/rad; $d_t\in[5,500]$\,Ns/m (up to $10^3$ on $t_x$); $d_r\in[0.3,15]$\,Nms/rad. These keep contact forces below half body weight and damping ratio $\zeta\in0.02$--$1.5$ at the HDT's pelvis inertia.}

\subsection{Statistical Characterization of Parameter Shareability}
\label{sec:shareability}
To assess whether the calibrated parameters generalize across users, we collect, for each of the $N=5$ calibration subjects, the $R=10$ fits $\hat{\theta}_{s,r}$ (each an instance of $\theta_{\mathrm{Opt}}$) produced by the dynamic identification of Sec.~\ref{sec:two-stage calibration} (five CMA-ES restarts on each of two dynamic-motion trials, $r=1,\dots,R$). Because the impedance parameters are strictly positive and span several orders of magnitude, a regime where multiplicative (log-normal) variation is more natural than additive Gaussian noise~\cite{Limpert2001}, the $N\cdot R=50$ fits per parameter are modeled in the same log-space $z=\ln\theta$ as the identification (Sec.~\ref{sec:two-stage calibration}), as a one-way random-effects decomposition
\begin{equation}
\ln\hat{\theta}_{s,r}=\mu+b_s+\epsilon_{s,r},\quad b_s\sim\mathcal{N}(0,\sigma_B^2),\quad \epsilon_{s,r}\sim\mathcal{N}(0,\sigma_W^2),
\label{eq:LMM}
\end{equation}
where $\mu$ is the population grand mean of the log-parameter, $b_s$ the between-subject (physiological) deviation, and $\epsilon_{s,r}$ the within-subject (restart-to-restart algorithmic) noise, with variances $\sigma_B^2$ and $\sigma_W^2$ respectively. Subject dependency is quantified by the single-measurement intraclass correlation coefficient (ICC), specifically $\mathrm{ICC}(1,1)$~\cite{KooLi2016}:
\begin{equation}
    \rho = \frac{\sigma_B^2}{\sigma_B^2 + \sigma_W^2},
\end{equation}
estimated by the method of moments from the one-way ANOVA within- and between-subject mean squares $MS_W,MS_B$, which are the sample estimators of the two variance components ($\mathbb{E}[MS_W]=\sigma_W^2$, $\mathbb{E}[MS_B]=\sigma_W^2+R\,\sigma_B^2$)~\cite{ShroutFleiss1979}; inverting these relations yields the closed form
\begin{equation*}
\hat{\rho}=\max\!\big(0,\,(MS_B-MS_W)/(MS_B+(R-1)MS_W)\big),
\end{equation*}
where the $\max(0,\cdot)$ enforces a non-negative variance and equivalently truncates a negative ICC at the boundary case $MS_W>MS_B$~\cite{Liljequist2019}. In our context, a low ICC means physiological differences are masked by optimizer noise, so the parameter is shareable across subjects~\cite{KooLi2016,PortneyWatkins2009}.

To handle the finite-sample uncertainty of $\hat{\rho}$ under $N=5$, we compute its $95\%$ confidence interval $[\rho_L,\rho_U]$ by a hierarchical (cluster) bootstrap that resamples whole subjects together with all $R$ of their fits~\cite{EfronTibshirani1994}. Following standard clinical reliability conventions~\cite{Cicchetti1994,PortneyWatkins2009}, each parameter is mapped to a deployable prior according to where its interval $[\rho_L,\rho_U]$ falls, as \emph{common} (i.e.\ shareable; $\mathcal{C}_{\mathrm{com}}$), \emph{distinct} ($\mathcal{C}_{\mathrm{dis}}$), or \emph{uncertain} ($\mathcal{C}_{\mathrm{unc}}$):
\begin{equation}
\theta_{\mathrm{prior}}=
\begin{cases}
\hat{\mu}, & \rho_U<0.50\;\;(\mathcal{C}_{\mathrm{com}}),\\
\text{per-subject calibration}, & \rho_L\ge0.75\;\;(\mathcal{C}_{\mathrm{dis}}),\\
[\hat{\mu}_L,\hat{\mu}_U], & \text{otherwise}\;\;(\mathcal{C}_{\mathrm{unc}}).
\end{cases}
\label{eq:prior_mapping}
\end{equation}
where the point prior $\hat{\mu}$ is the geometric mean of the 50 pooled fits ($\exp$ of the log grand mean $\mu$ in~\eqref{eq:LMM}) and $[\hat{\mu}_L,\hat{\mu}_U]$ their empirical $95\%$ spread (the 2.5th and 97.5th percentiles of those fits). For uncertain parameters, this spread provides a bounded search range when configuring an unseen user.

To verify that the prior derived from~\eqref{eq:prior_mapping} actually transfers to an unseen user, we evaluate it under a leave-one-subject-out (LOSO) protocol. For each held-out subject $s$, the prior $\theta_{\mathrm{prior}}^{(-s)}$ is the per-parameter geometric mean of the remaining $N-1$ subjects' pooled fits ($R=10$ restarts each), and is simulated with the subject's own calibrated joint limits, so the residual cost reflects only the impedance prior. The per-subject overhead, defined as the relative increase of cost $J$ or position RMSE over that subject's own best fit,
\begin{equation}
\Delta^{(s)}_M = \frac{M(\theta_{\mathrm{prior}}^{(-s)})-M(\hat{\theta}_s^{*})}{M(\hat{\theta}_s^{*})},\quad M\in\{J,\,\mathrm{RMSE}_{\mathrm{pos}}\},
\label{eq:loso_overhead}
\end{equation}
averaged across the cohort, then measures how much fidelity is lost when the shared prior replaces a per-subject calibration. A small $\bar{\Delta}$ is taken as empirical evidence that the shareability classification and the resulting prior in~\eqref{eq:prior_mapping} transfer to unseen users.\footnote{In this study we deploy $\theta_{\mathrm{prior}}^{(-s)}$ as a single point estimate across all 12 parameters, i.e.\ the uncertain branch ($\mathcal{C}_{\mathrm{unc}}$) of~\eqref{eq:prior_mapping} is collapsed to its point prior $\hat{\mu}$. This is a conservative baseline; interval-based search for $\mathcal{C}_{\mathrm{unc}}$ parameters is left as future work once a larger cohort sharpens the bounds.}

\subsection{Model Validation Strategy}\label{sec:validation_strategy}

{With the prior derivation already validated under LOSO (Sec.~\ref{sec:shareability}), we now assess the calibrated model itself along two complementary axes: the fidelity of the interaction envelope it produces on an unseen subject, and whether this spatial accuracy propagates into correct downstream biomechanical adaptation.}

\paragraph{Interaction envelope} To verify that the calibrated model accurately replicates real-world physical interaction boundaries, we evaluate the interface midpoint position relative to the pelvis center, $\mathbf{r}=\mathbf{p}_{\mathrm{mid}}-\mathbf{p}_{\mathrm{pelvis}}$; this subtraction removes the overall body movement common to both bodies and isolates the residual strap--pelvis compliance, which is precisely what the viscoelastic model governs. Following standard postural-sway characterization~\cite{Prieto1996,Winter1995}, the resulting 3-D cloud of $\mathbf{r}$ is summarized by its per-axis marginal spreads $\boldsymbol{\sigma}=(\sigma_x,\sigma_y,\sigma_z)$ (forward, lateral, and vertical) and its transverse-plane anisotropy ratio $\gamma=\sigma_x/\sigma_y$. Because natural human pelvic sway is characteristically laterally dominant ($\gamma<1$, whereas $\gamma\to1$ signals an unrealistic isotropic cloud), a physically accurate model must simultaneously recover the absolute spatial scale ($\boldsymbol{\sigma}$) and the true directional bias ($\gamma$) of the real motion.

\paragraph{Biomechanical adaptation} Finally, we test whether the HDT autonomously replicates the gait adaptations of real robot-assisted walking. Sagittal-plane hip/knee/ankle trajectories are segmented by heel strike, time-normalized to $0$--$100\%$ gait cycle, and averaged (mean$\pm$SD) over 8 cycles per condition. Agreement with the real reference is assessed at three complementary levels: Range of Motion (RoM), peak joint angle, and waveform similarity (Pearson $r$). The clinically relevant signature is the robot-induced reduction in lower-limb RoM: the strap constrains the pelvis and, via the pelvis$\to$hip$\to$knee$\to$ankle chain, propagates to all three joints. Only a correctly identified model should induce this reduction, bracketed by overly compliant (near-unconstrained) and overly stiff (excessively suppressed) failure modes that manual tuning cannot reliably distinguish.

\section{EXPERIMENTS}
\label{sec4:experiment}
The experimental protocol involving human subjects was approved by the Institutional Review Board (IRB) of Nanyang Technological University (Approval Number: NTU IRB-2024-257 and NTU IRB-2025-076).

\begin{figure}[htbp]
    \centering
    \includegraphics[width=\linewidth]{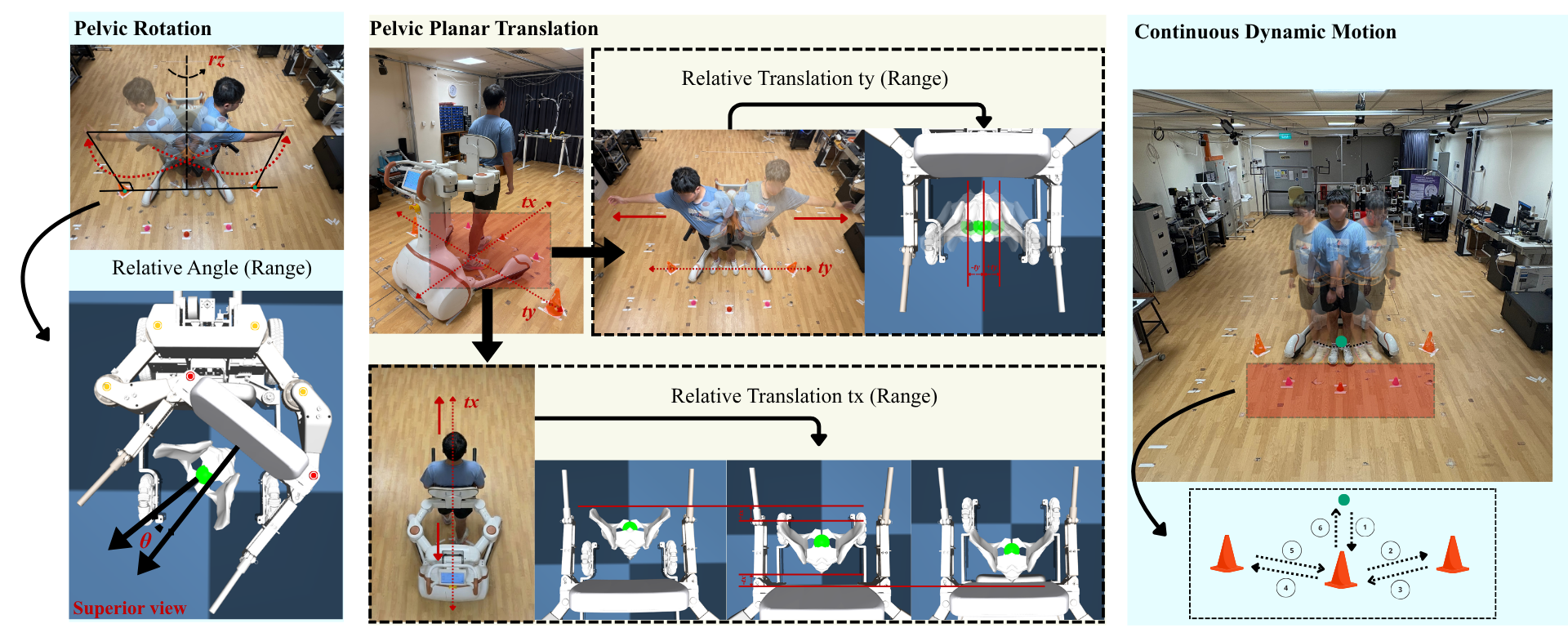}
    \caption{Calibration Tasks. \textbf{Left:} Maximum pelvic rotation. \textbf{Center:} Maximum planar translation. \textbf{Right:} Continuous dynamic motion for viscoelastic excitation. \textbf{Bottom-left (superior view):} defines the relative pelvic rotation angle $\Delta\phi$ (pelvis vs.\ interface) and shows the markers tracked by the Qualisys motion-capture system (red and yellow); the two red markers lie at the interface joint centers, whose midpoint defines the interface midpoint.}
    \label{fig:calibration_motions}
\end{figure}

To evaluate the efficacy of the proposed Real2Sim calibration and validation framework, we conducted a pilot study involving six healthy subjects. The participants were divided into two cohorts: a \textit{calibration cohort} ($N=5$) used to derive the parameter statistics, and an \textit{unseen validation subject} ($N=1$) used to test the pipeline's generalizability. Participants were recruited to represent a diverse range of anthropometric profiles, particularly regarding waist circumference (Age: $21\text{--}40$ years; Mass: $82.7\pm10.4$ kg; Height: $174.7\pm4.2$ cm; Waist: $92.3\pm10.2$ cm). 

Data collection used a hybrid setup. A markerless system~\cite{jatesiktat2024anatomical} reconstructs a per-subject scaled Rajagopal OpenSim model~\cite{rajagopal2016fullbody}, yielding the pelvis 6-DoF pose and hip/knee/ankle angles for biomechanical validation; a marker-based system (Qualisys) tracks the two interface joint centers and the pelvic ASIS/PSIS markers (Fig.~\ref{fig:calibration_motions}). The two were spatially aligned and temporally synchronized.

\subsection{pHRI Model Parameter Calibration and Identification}

\subsubsection{Real-world Procedure}
Subjects performed a standardized calibration protocol across multiple discrete tightness levels (8--10 per subject, spanning $[T_{\min}, T_{\max}]$ in 1-cm increments). At each level they completed three tasks (Fig.~\ref{fig:calibration_motions}): \emph{pelvic rotation} to its angular limits via contralateral reaching; \emph{planar translation} to lateral limits (ipsilateral reaching) and antero-posterior limits (``leaning against'' / ``dragging'' the robot); and \emph{continuous dynamic motion} toward three target cones to excite the viscoelastic dynamics. The first two identify the joint limits ($\mathbf{q}_{\min},\mathbf{q}_{\max}$); the third excites the impedance.
To standardize execution speed, participants practiced with a 100\,bpm metronome before recording and then performed at this self-maintained pace. After all trials, each subject identified their ``Safe \& Comfortable'' operating point $T^*$ (Sec.~\ref{sec:tightness}).

\subsubsection{Simulation Procedure}
Parameter identification requires the simulation to replicate the exact user kinematics observed in reality. We implemented a kinematic replay approach where a \texttt{mocap} body attached to the digital twin’s pelvis strictly tracks the recorded real-world trajectory frame-by-frame. This forces the digital twin to replicate the user’s
exact motion, treating the human kinematics as a fixed input that drives the interaction, rather than a dynamic body that reacts to forces.
The stiffness and damping parameters were then identified with CMA-ES~\cite{hansen2016cma} in the log-space reparameterization $z=\ln\theta$ of Sec.~\ref{sec:two-stage calibration}.\footnote{CMA-ES: box bounds $[\ln\theta_{\min},\ln\theta_{\max}]$, $\sigma_0\approx1.22$, $\lambda=11$, $1200$ evaluations, $\mathrm{tolfun}=10^{-6}$. Each subject's two dynamic trials are each optimized with $5$ log-uniform restarts ($10$ fits/subject): their scatter feeds the variance decomposition of Sec.~\ref{sec:shareability} ($R=10$), and each trial's best fit is forward-simulated on the other for cross-trial generalization (Sec.~\ref{sec:rq2_results}).}
\subsection{pHRI Model Validation}
\subsubsection{Real-world Procedure}
The participant performed Normal Walking (NW) after pre-training with a 100\,bpm metronome for a steady cadence. This NW dataset served as the kinematic baseline to train the HDT. The participant then performed robot-assisted walking at their self-selected $T^*$ tightness. To isolate physical adaptations from cognitive safety strategies, the robot's maximum velocity was capped slightly below the user's natural speed. This five-trial session provided both the full-body kinematics for gait analysis and the marker-based interface midpoint as the real-world pHRI reference.

\subsubsection{Simulation Procedure}

The HDT was trained on the baseline NW motions with domain randomization (stochastic external forces $[-100,100]$\,N at random gait phases) for disturbance rejection, then evaluated in coupled simulation with the robot configured identically to the experiment under three pHRI conditions (soft, $\theta_{\mathrm{prior}}$, stiff). For each, simulated interface-midpoint trajectories relative to the pelvis (Sec.~\ref{sec:validation_strategy}; low-pass filtered at 12\,Hz) and downstream sagittal-plane kinematics were compared against the real reference.

\section{RESULTS AND DISCUSSION}

\subsection{Analyzing the Safe-Comfortable Tightness Threshold}

\begin{figure}[htbp]
    \centering
    \includegraphics[width=1\linewidth]{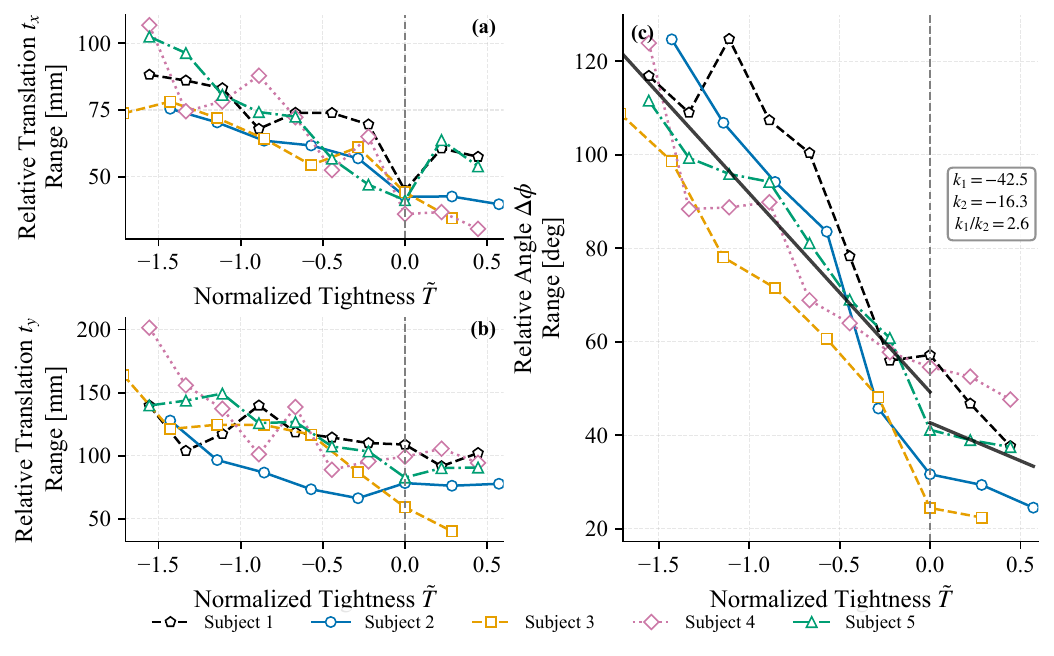}
    \caption{Cross-subject tightness sweep of allowable relative pelvis--interface motion against normalized strap tightness $\tilde{T}$ (Eq.~\ref{eq:method4}; centered at each subject's Safe \& Comfortable level $T^*$). (a, b) Maximum allowable planar translation ranges $\Delta x$ and $\Delta y$ (mm) along the antero-posterior and medio-lateral axes show no consistent cross-subject transition. (c) Maximum allowable pelvic rotation range $\Delta\phi$ (deg) about the vertical axis ($r_z$ in our model); a clear knee point (rapid drop then plateau) emerges across subjects. Vertical markers denote each subject's self-selected Safe \& Comfortable tightness, which coincides with the knee point in (c). The two grey diagonal lines in (c) are piecewise linear regressions on the pooled left/right data; $k_1$ and $k_2$ are the corresponding slopes.}
    \label{fig:motion5_distance_comparison}
\end{figure}

To select a representative coupling condition, we evaluated how allowable relative motion varies with strap tightness using $\Delta x$, $\Delta y$, and pelvic rotation $\Delta\phi$ (Fig.~\ref{fig:motion5_distance_comparison}). While $\Delta x$ and $\Delta y$ show no consistent cross-subject transition, $\Delta\phi$ exhibits a clear knee point (rapid drop then plateau), marking a shift from slack to constrained coupling. A two-segment piecewise linear regression on each subject's pooled left/right data shows a sharp change in slope between the two fitted segments ($k_1$ vs.\ $k_2$), indicating a breakpoint. It coincides with the subject's self-selected ``Safe \& Comfortable'' tightness (vertical markers). The same fit applied to $\Delta x$ and $\Delta y$ yields no shared breakpoint. Any apparent feature (e.g.\ the drop-then-plateau of $\Delta x$ near level~7) appears in isolated subjects only, in neither consistent location nor slope sign, so planar slack is not a reliable operating-point indicator. Beyond the knee point, tighter settings add negligible restriction while risking discomfort; we therefore adopt the knee-point tightness as each subject's operating point for the subsequent identification, simulation, and experiments (\textbf{RQ1}).
\subsection{Optimization Efficacy and Parameter Classification}
\label{sec:rq2_results}
\subsubsection{Identification Efficacy and Cross-Trial Generalization}

The CMA-ES pipeline converges robustly: for every subject the five restarts collapse onto a consistent plateau of the dimensionless cost $J$ (Fig.~\ref{fig:classification}, top), insensitive to initialization despite the non-convex landscape.

To verify that the identified parameters reflect the underlying interface impedance rather than overfitting trial-specific noise, we cross-validate each subject's two continuous-dynamic-motion trials. The best-of-five fit from one trial is forward-simulated on the held-out trial, and the two train/test directions are averaged to remove asymmetry. Across the five subjects the cross-trial cost averages $J=1.46$, a modest $+14.5\%$ overhead above the within-trial best fit ($J=1.28$, per-subject overhead $+4.7\%$ to $+25.2\%$). The consistency between within- and cross-trial fits indicates that the calibrated viscoelastic parameters capture a stable, trial-invariant compliance of the pelvis--strap interface.

\subsubsection{Statistical Characterization of Shareability}

\begin{figure}[htbp]
    \centering
    \includegraphics[width=\linewidth]{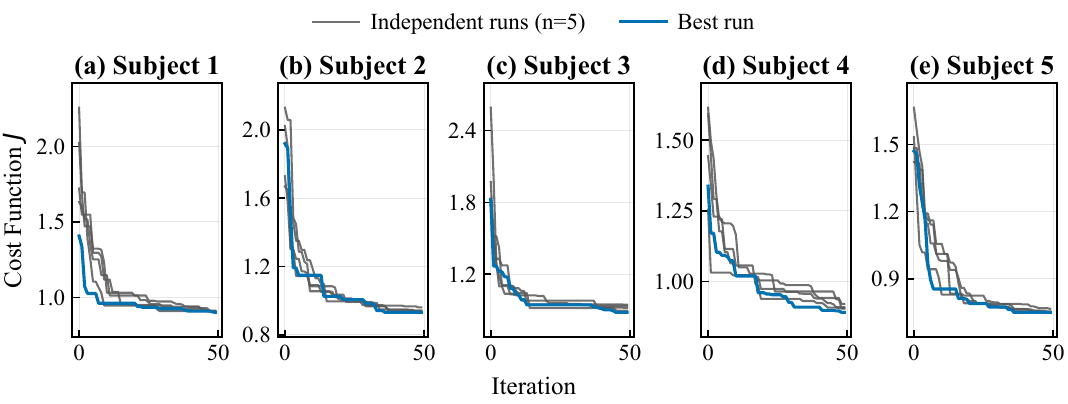}\\[2pt]
    \includegraphics[width=\linewidth]{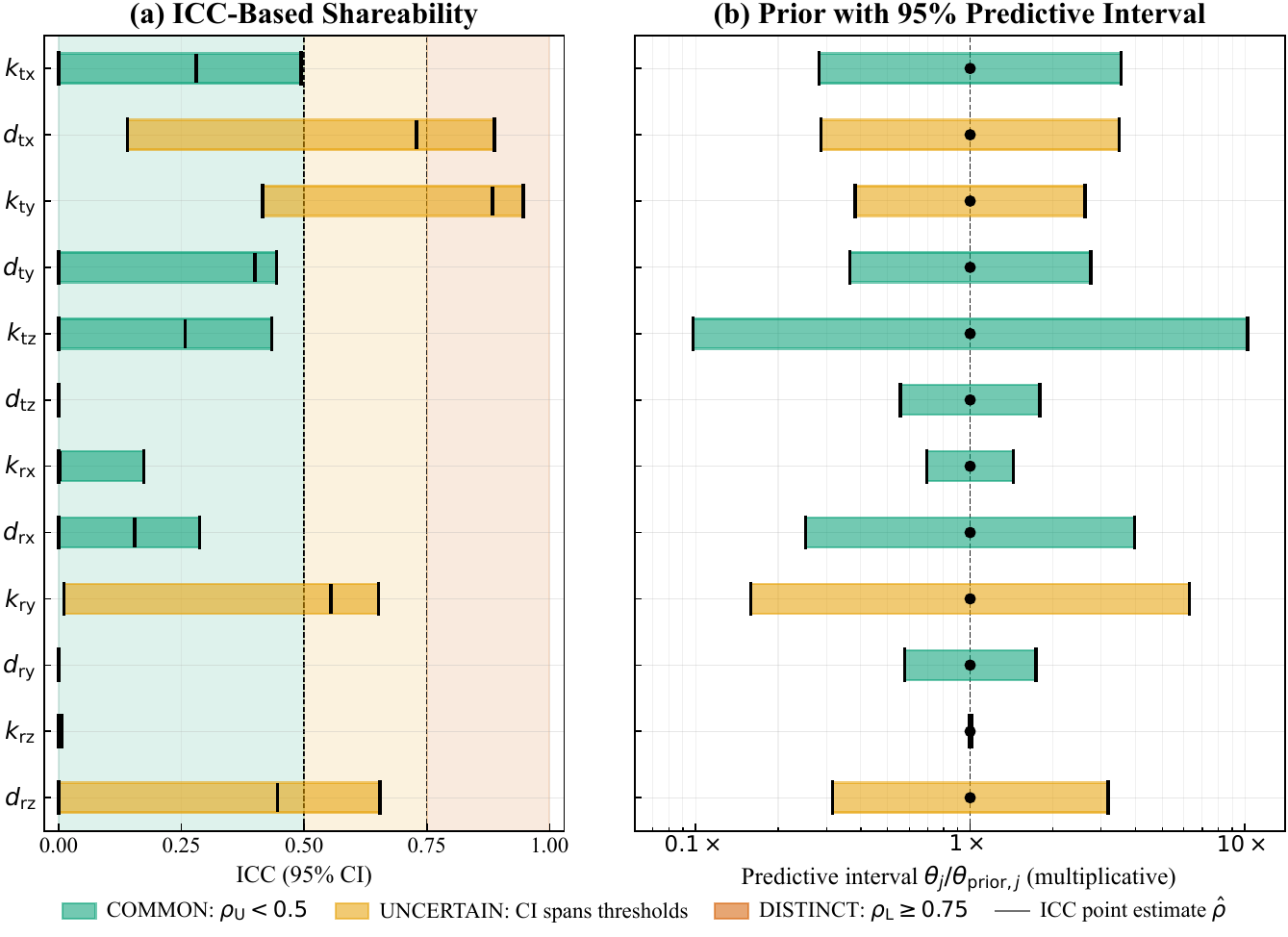}\\[2pt]
    \includegraphics[width=\linewidth]{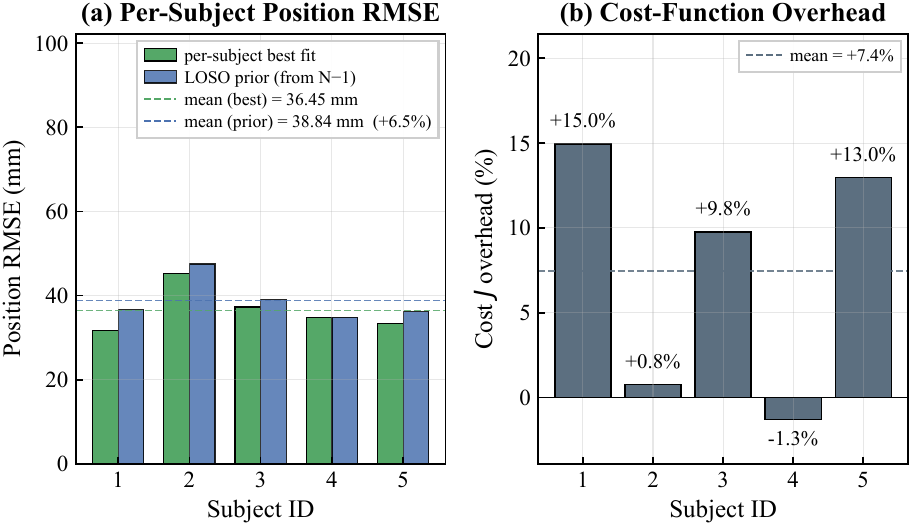}
    \caption{Identification and generalization diagnostics. \textbf{Top:} CMA-ES convergence, dimensionless cost $J$ vs.\ iteration per subject (grey: five restarts, blue: best run), collapsing onto a consistent plateau despite the non-convex landscape. \textbf{Middle:} ICC-based shareability: (a) ICC with bootstrap 95\% CI; (b) parameter distribution. \textbf{Bottom:} leave-one-subject-out (LOSO) generalization cost, per-subject position RMSE and cost-$J$ overhead of the shared prior relative to the per-subject best fit (mean $+6.5\%$ RMSE, $+7.4\%$ $J$); Subject~4's slight negative overhead is a harmless artifact of the flat cost landscape.}
    \label{fig:classification}
\end{figure}

We utilized the ICC to distinguish true physiological differences from optimizer noise, answering \textbf{RQ2}:

\paragraph{Shareable parameters ($\mathcal{C}_{\mathrm{com}}$)}
Seven of the twelve parameters fell into the \textsc{shareable} category, with negligible subject dependency ($\rho_U < 0.50$; green zone of Fig.~\ref{fig:classification}(a)): the translational terms $k_{tx}, d_{ty}, k_{tz}, d_{tz}$ and the rotational terms $k_{rx}, d_{ry}, k_{rz}$. These cluster tightly around the population mean (Fig.~\ref{fig:classification}(b)), dominated by invariant harness material properties rather than individual physiology, and are deployed as fixed point priors (Table~\ref{tab:shared_priors_simplified}). Notably, the torsional $k_{rz}$ prior ($15.2$\,Nm/rad) sits at its lower admissible bound, so its \textsc{shareable} classification reflects boundary saturation rather than a genuine physiological consensus.

\paragraph{Subject-specific parameters ($\mathcal{C}_{\mathrm{unc}}$)}
The remaining five parameters, $d_{tx}, k_{ty}$ (translational) and $d_{rx}, k_{ry}, d_{rz}$ (rotational), fell into the \textsc{uncertain} category, exhibiting markedly higher between-subject variance (Fig.~\ref{fig:classification}(b)). For these a single fixed value is inadequate; instead we deploy the $95\%$ prediction interval (PI) of Table~\ref{tab:shared_priors_simplified} as a bounded search range, so an unseen user is configured by refining only these dimensions, reducing per-user tuning from 12 to 5 parameters while reaching high fidelity (\textbf{RQ2}). The present cohort ($N=5$, two trials each) is too small to regress these values on anthropometrics (e.g., waist, mass), so we report only the population point prior and its PI and leave subject-level regression to a larger cohort.

\begin{table}[htbp]
\caption{Recommended shared priors. For each parameter, $\theta_{\mathrm{prior}}$ is the population point estimate and the 95\% prediction interval (PI) its bounded search range. \textsc{shareable} parameters are deployed directly at $\theta_{\mathrm{prior}}$; \textsc{uncertain} parameters are refined within the PI.}
\label{tab:shared_priors_simplified}
\centering
\renewcommand{\arraystretch}{0.95}
\setlength{\tabcolsep}{4pt}
\resizebox{\columnwidth}{!}{%
\begin{tabular}{lrccc}
\toprule
\textbf{Param.} & \boldmath$\theta_{\mathrm{prior}}$ & \textbf{95\% PI} & \textbf{Unit} & \textbf{ICC class} \\
\midrule
\multicolumn{5}{l}{\textit{Translational Impedance}} \\
$k_{tx}$ & 725  & [205,\,2570]   & N/m   & \textsc{shareable} \\
$d_{tx}$ & 588  & [169,\,2050]   & Ns/m  & \textsc{uncertain} \\
$k_{ty}$ & 2120 & [808,\,5540]   & N/m   & \textsc{uncertain} \\
$d_{ty}$ & 133  & [48.3,\,364]   & Ns/m  & \textsc{shareable} \\
$k_{tz}$ & 3850 & [377,\,39400]  & N/m   & \textsc{shareable} \\
$d_{tz}$ & 70.8 & [39.4,\,127]   & Ns/m  & \textsc{shareable} \\
\midrule
\multicolumn{5}{l}{\textit{Rotational Impedance}} \\
$k_{rx}$ & 39.2 & [27.3,\,56.3]  & Nm/rad  & \textsc{shareable} \\
$d_{rx}$ & 1.26 & [0.318,\,5.02] & Nms/rad & \textsc{uncertain} \\
$k_{ry}$ & 122  & [19.5,\,768]   & Nm/rad  & \textsc{uncertain} \\
$d_{ry}$ & 2.20 & [1.27,\,3.82]  & Nms/rad & \textsc{shareable} \\
$k_{rz}$ & 15.2 & [15.0,\,15.3]  & Nm/rad  & \textsc{shareable} \\
$d_{rz}$ & 7.21 & [2.27,\,22.9]  & Nms/rad & \textsc{uncertain} \\
\bottomrule
\end{tabular}}
\end{table}

\paragraph{Generalization to unseen users (LOSO)}
We next test whether the shared prior transfers to a held-out user under the leave-one-subject-out protocol of Eq.~\eqref{eq:loso_overhead}. Deploying each subject's leave-one-out prior on that subject's own model incurs only a small overhead over their per-subject best fit: on average $+6.5\%$ in position RMSE and $+7.4\%$ in cost $J$ (Fig.~\ref{fig:classification}, bottom). Subject~4 shows a slight negative overhead, a harmless artifact of the flat cost landscape near the optimum. This small overhead confirms that the shareability classification and the resulting prior generalize to unseen users with minimal loss of fidelity (\textbf{RQ2}).

\subsection{pHRI Validation}
We validate the calibrated coupling model during dynamic overground walking along two complementary axes: the \textit{interaction envelope} at the physical interface and downstream \textit{biomechanical adaptation} of the user's gait. The deployed population prior ($\theta_{\mathrm{prior}}$) is compared against two bounding isotropic baselines: an overly compliant setting (\textit{soft}: $\mathbf{K}=10\mathbf{I}$, $\mathbf{D}=\mathbf{I}$) and an overly rigid setting (\textit{stiff}: $\mathbf{K}=10^4\mathbf{I}$, $\mathbf{D}=10^3\mathbf{I}$).

\subsubsection{Interaction envelope}
\begin{figure}[htbp]
    \centering
    \includegraphics[width=\linewidth]{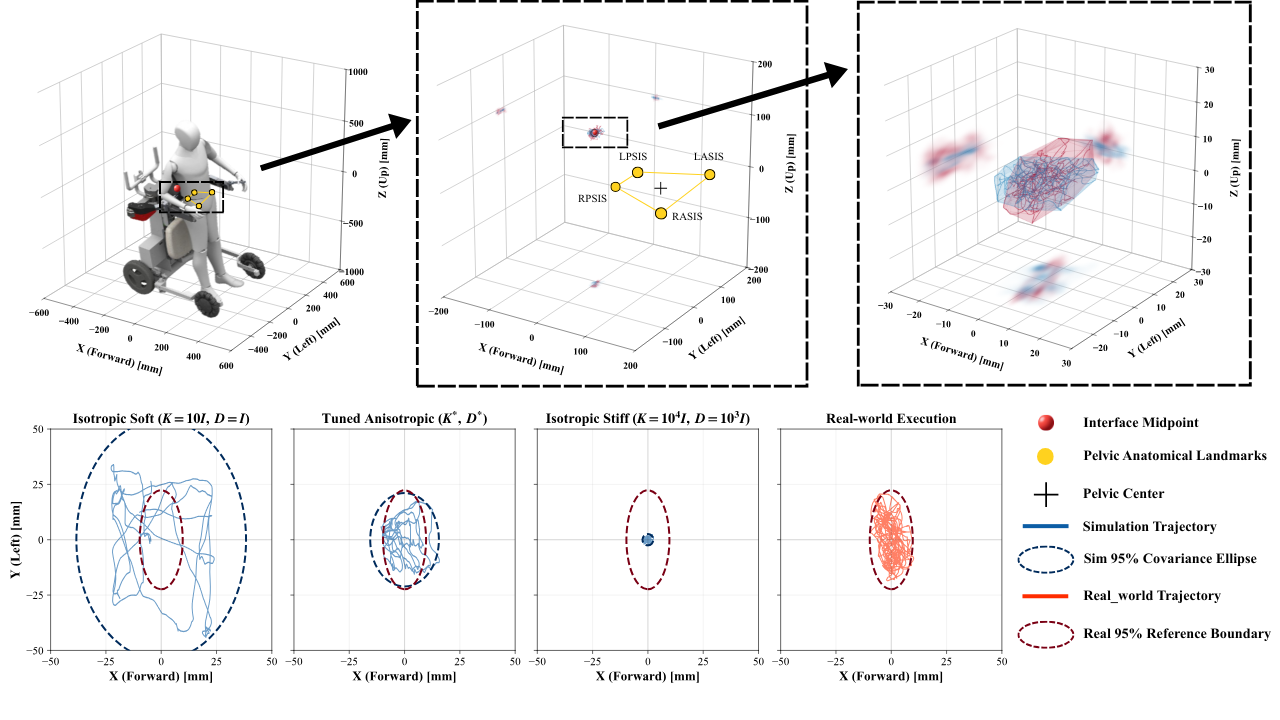}
    \caption{Spatial interaction validation during robot-assisted walking. \textbf{Top:} Geometric hierarchy mapping the full robot-human model down to the marker-level pelvis landmarks and the interface midpoint tracker. \textbf{Bottom:} Transverse plane ($X$--$Y$) interface displacement trajectories and corresponding $95\%$ confidence ellipses. The baseline settings highlight clear modeling limits: the \textit{soft} interface allows extensive over-travel, while the \textit{stiff} interface collapses natural sway. The calibrated parameters $\theta_{\mathrm{prior}}$ mirror the true directional boundaries and shape of the real physical interaction.}
    \label{fig:drba_validation}
\end{figure}

Fig.~\ref{fig:drba_validation} and Table~\ref{tab:interface_dist} examine the spatial structure of the pelvis--interface displacement to test whether the calibrated model replicates the real interaction boundaries.

Improper interface tuning is exposed by the $X$--$Y$ trajectory clouds and $95\%$ confidence ellipses in Fig.~\ref{fig:drba_validation}. The baseline settings illustrate clear failure modes: the \textit{soft} interface under-constrains the pelvis and significantly over-travels beyond the real interaction envelope, while the \textit{stiff} interface over-constrains the user, collapsing natural pelvic movement into a near-static region. The calibrated parameter set $\theta_{\mathrm{prior}}$ tightly captures the size, orientation, and boundary shape of the real interaction.

Quantitatively, Table~\ref{tab:interface_dist} tracking postural sway characteristics corroborates these trends. The \textit{soft} configuration allows excessive lateral drift ($\sigma_y = 22.07$~mm), while the \textit{stiff} configuration artificially flattens the trajectory into symmetric, isotropic behavior ($\gamma = 0.951$). Only $\theta_{\mathrm{prior}}$ successfully balances these extremes, accurately recovering both the true physical scale ($\boldsymbol{\sigma} = [3.04, 8.44, 2.21]$~mm vs.\ real $[3.82, 9.81, 4.19]$~mm) and the laterally-dominant anisotropy ($\gamma = 0.360$ vs.\ real $0.389$) characteristic of natural pelvic sway. Capturing directional anisotropy, not assuming uniform isotropic constants, is thus essential to physical interaction fidelity.

\begin{table}[htbp]
\caption{Postural sway characterization at the physical interface. Marginal spatial spreads $\sigma_x$ (forward), $\sigma_y$ (lateral), and $\sigma_z$ (vertical) are compared alongside the in-plane anisotropy ratio $\gamma=\sigma_x/\sigma_y$ for an unseen subject across five experimental trials. Natural human pelvic sway is laterally dominant ($\gamma < 1$); the calibrated parameters $\theta_{\mathrm{prior}}$ uniquely match both the absolute spatial scale and the directional bias of the real-world reference.}
\label{tab:interface_dist}
\centering
\renewcommand{\arraystretch}{1.1}
\setlength{\tabcolsep}{6pt}
\footnotesize
\begin{tabular}{lllcccc}
\toprule
 & \multicolumn{2}{l}{\textbf{Settings}} & \boldmath$\sigma_x$ & \boldmath$\sigma_y$ & \boldmath$\sigma_z$ & \boldmath$\gamma$ \\
 & & & (mm) & (mm) & (mm) & $=\sigma_x/\sigma_y$ \\
\midrule
\multirow{3}{*}{\textbf{Sim}} & \multicolumn{2}{l}{Soft ($k=10$)}        & 15.56 & 22.07 & 2.43 & 0.705 \\
 & \multicolumn{2}{l}{$\theta_{\mathrm{prior}}$} & \textbf{3.04}  & \textbf{8.44}  & 2.21 & \textbf{0.360} \\
 & \multicolumn{2}{l}{Stiff ($k=10^4$)}     & 0.97  & 1.02  & 3.28 & 0.951 \\
\cmidrule{1-7}
\textbf{Real} & \multicolumn{2}{l}{}  & 3.82  & 9.81  & 4.19 & 0.389 \\
\bottomrule
\end{tabular}
\end{table}

\subsubsection{Biomechanical adaptation}
\begin{figure}[htbp]
    \centering
    \includegraphics[width=\linewidth]{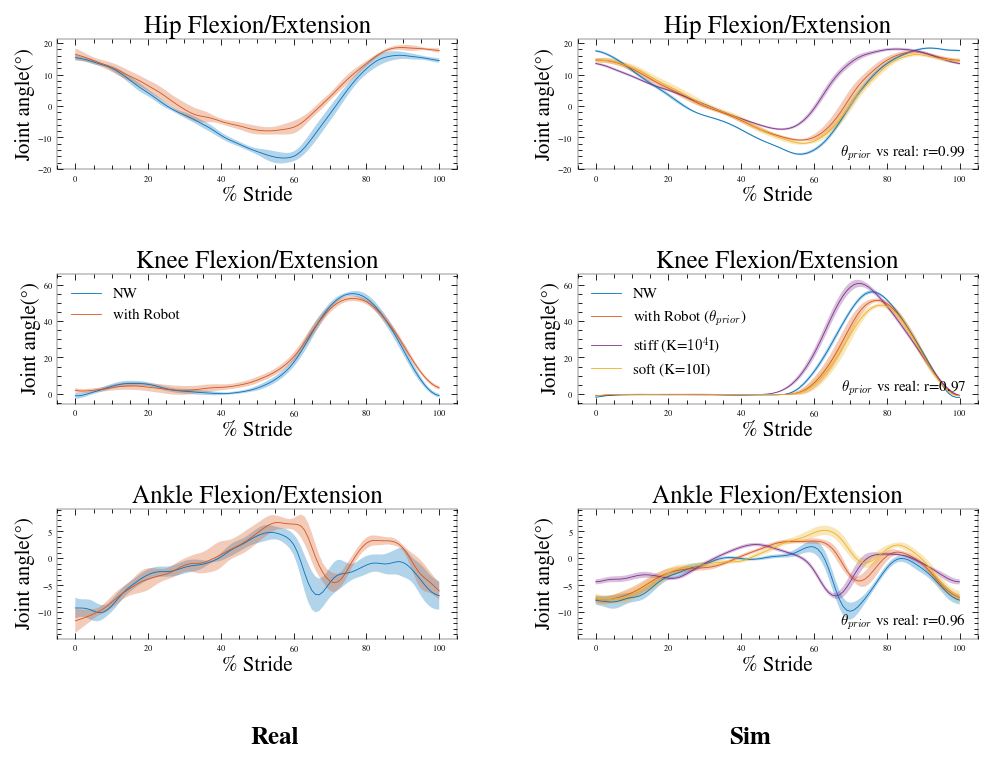}
    \caption{Sagittal-plane joint kinematics for gait validation. \textbf{Left (Real)}: Measured joint angles during Normal Walking (NW) and robot-assisted walking with the DRBA. \textbf{Right (Sim)}: Simulated joint angles of the HDT under NW and under the robot with three impedance settings: the deployed prior ($\theta_{\mathrm{prior}}$), an overly compliant isotropic interface (\emph{soft}, $\mathbf{K}=10\mathbf{I},\,\mathbf{D}=\mathbf{I}$), and an overly stiff isotropic interface (\emph{stiff}, $\mathbf{K}=10^4\mathbf{I},\,\mathbf{D}=10^3\mathbf{I}$). The shaded areas represent the standard deviation of the joint angles across multiple gait cycles.}
    \label{fig:joint_kinematics}
\end{figure}

As shown in Fig.~\ref{fig:joint_kinematics} and Table~\ref{tab:stm002_kinematics}, we evaluate if the identified interface parameters allow the digital twin to autonomously replicate the downstream kinematic adaptations of real robot-assisted gait. In the physical experiment, transitioning from Normal Walking (NW) to robot-assisted walking compresses lower-limb sagittal Range of Motion (RoM), dropping hip RoM by $6.4^\circ$ ($33.4^\circ\to27.0^\circ$) and knee RoM by $5.4^\circ$ ($57.5^\circ\to52.1^\circ$). This is driven primarily by a truncated peak hip extension and reduced ankle dorsiflexion before toe-off ($\approx$60\% of the gait cycle) as the physical harness limits stride-length energy transfer. To isolate the physical validity of the pHRI model from cognitive safety strategies, the robot's maximum velocity was capped slightly below the user's natural speed. This ensures observed adaptations are driven by physical interaction dynamics rather than voluntary psychological slowing.

Deploying the population prior $\theta_{\mathrm{prior}}$ on this unseen subject, the unguided HDT successfully reproduces this natural human adaptation at the RoM level without any compensatory retraining. At the hip, the simulated reduction drops by $-5.5^\circ$ (vs.\ real $-6.4^\circ$), tracking the overall trajectory profile closely ($r=0.99$). At the knee, the restriction matches the physical benchmark exactly, yielding an emergent reduction of $-5.4^\circ$ (vs.\ real $-5.4^\circ$). This high-precision tracking demonstrates that the calibrated stiffness and damping matrices accurately model the "hard stop" mechanics of the strap, forcing the digital twin to correctly adapt to physical boundary conditions (\textbf{RQ3}).

The primary differences are localized to the ankle. The simulated unassisted ankle NW baseline is initially underestimated ($12.1^\circ$ vs.\ $16.9^\circ$), a discrepancy common in HDT where the policy develops alternative stabilization strategies to handle complex contact conditions. Despite this baseline offset, the characteristic decrease in ankle dorsiflexion before toe-off remains clearly observable in both simulation and reality when walking coupled to the DRBA.

\begin{table}[t]
\centering
\caption{Gait kinematics of the unseen subject: range of motion (RoM) and peak angle of each condition. Sim conditions combine a fixed NW-style policy with the robot; $\theta_{\mathrm{prior}}$ uses the population-prior interface impedance.}
\label{tab:stm002_kinematics}
\setlength{\tabcolsep}{4pt}
\renewcommand{\arraystretch}{0.9}
\scriptsize
\begin{tabular}{llcc}
\toprule
Joint & Condition & RoM ($^\circ$) & Peak ($^\circ$) \\
\midrule
\multirow{6}{*}{Hip} & NW (real) & 33.4$\pm$1.8 & 16.7 \\
 & with Robot (real) & 27.0$\pm$1.6 & 18.9 \\
 & NW (sim) & 33.8$\pm$0.5 & 18.4 \\
 & $\theta_{\mathrm{prior}}$ (sim) & 28.3$\pm$0.5 & 17.3 \\
 & stiff (sim) & 25.6$\pm$0.6 & 18.2 \\
 & soft (sim) & 28.5$\pm$0.6 & 16.4 \\
\midrule
\multirow{6}{*}{Knee} & NW (real) & 57.5$\pm$2.0 & 55.5 \\
 & with Robot (real) & 52.1$\pm$1.2 & 52.6 \\
 & NW (sim) & 58.4$\pm$0.8 & 56.3 \\
 & $\theta_{\mathrm{prior}}$ (sim) & 53.0$\pm$1.0 & 51.8 \\
 & stiff (sim) & 62.2$\pm$1.7 & 61.1 \\
 & soft (sim) & 50.5$\pm$1.1 & 49.2 \\
\midrule
\multirow{6}{*}{Ankle} & NW (real) & 16.9$\pm$1.4 & 5.4 \\
 & with Robot (real) & 19.6$\pm$1.8 & 7.6 \\
 & NW (sim) & 12.1$\pm$0.6 & 2.2 \\
 & $\theta_{\mathrm{prior}}$ (sim) & 11.3$\pm$0.7 & 3.3 \\
 & stiff (sim) & 9.6$\pm$0.8 & 2.5 \\
 & soft (sim) & 13.3$\pm$1.2 & 5.3 \\
\bottomrule
\end{tabular}
\end{table}

\section{CONCLUSIONS AND FUTURE WORK}

We presented a Real2Sim pipeline that identifies, rather than hand-tunes, the 6-DoF viscoelastic dynamics of a pelvis--strap interface for gait-assistive Human-in-the-Loop simulation. A ``Safe \& Comfortable'' tightness anchors a reproducible operating point; CMA-ES and ICC analysis then yield seven shareable priors and five subject-specific parameters, configuring an unseen user by tuning only five. On that subject the model recovers the real interface displacement and reproduces the sagittal RoM reduction, outperforming the bracketing soft and stiff baselines, and thereby turns the HDT into a predictive tool for pre-clinical verification of personalized controllers.
\textbf{Future work} will add interval-based prior search, a closed-loop robot-aware HDT, force-level interface validation, and broader and impaired populations.


\addtolength{\textheight}{-1cm}   


\bibliographystyle{IEEEtran}
\bibliography{pHRI}
\end{document}